\def\year{2020}\relax
\documentclass[letterpaper]{article} 
\usepackage{aaai20}  
\usepackage{times}  
\usepackage{helvet} 
\usepackage{courier}  
\usepackage[hyphens]{url}  
\usepackage{graphicx} 
\usepackage{graphicx}  
\usepackage{multirow}
\usepackage{setspace}
\usepackage{tikz-dependency}
\usepackage{amsmath}
\usepackage{subfigure}

\usepackage{array}
\newcommand{\PreserveBackslash}[1]{\let\temp=\\#1\let\\=\temp}
\newcolumntype{C}[1]{>{\PreserveBackslash\centering}p{#1}}
\newcolumntype{R}[1]{>{\PreserveBackslash\raggedleft}p{#1}}
\newcolumntype{L}[1]{>{\PreserveBackslash\raggedright}p{#1}}
\makeatletter
\def\blfootnote{\xdef\@thefnmark{}\@footnotetext}
\makeatother

  \newcommand{\citet}[1]
  {\citeauthor{#1} ̃\shortcite{#1}}
  \newcommand{\citep}{\cite}

\title{Multi-Task Self-Supervised  Learning for Disfluency Detection}

\author{Shaolei Wang$^{1}$, Wanxiang Che$^{1}$\thanks{Corresponding Author} , Qi Liu$^{2}$, Pengda Qin$^{3}$,  Ting Liu$^{1}$, William Yang Wang$^{4}$  \\
	$^{1}$Center for Social Computing and Information Retrieval, Harbin Institute of Technology, China  \\
	$^{2}$University of Oxford\\
	$^{3}$Beijing University of Posts and Telecommunications, China\\
	$^{4}$University of California, Santa Barbara, CA, USA\\
	\{slwang, car, tliu\}@ir.hit.edu.cn, qi.liu@st-hughs.ox.ac.uk, qinpengda@bupt.edu.cn, william@cs.ucsb.edu\\	
}

\begin{document}

\maketitle
\begin{abstract}
	Most existing approaches to disfluency detection heavily rely on human-annotated data, which is expensive to obtain in practice. 
	To tackle the training data bottleneck, we investigate methods for combining multiple self-supervised tasks-i.e., supervised tasks where data can be collected without manual labeling.
	First, we construct large-scale pseudo training data by randomly adding or deleting words from unlabeled news data, and propose two self-supervised pre-training tasks:
	(i) tagging task to detect the added noisy words.
	(ii) sentence classification to distinguish original sentences from grammatically-incorrect sentences.
	We then combine these two tasks to jointly train a network.
	The pre-trained network is then fine-tuned using human-annotated disfluency detection training data. 
	Experimental results on the commonly used English Switchboard test set show that our approach can achieve competitive performance compared to the previous systems (trained using the full dataset) by using less than 1\% (1000 sentences) of the training data. 
	Our method trained on the full dataset significantly outperforms previous methods, reducing the error by 21\% on English Switchboard.
\end{abstract}

\section{Introduction}

\begin{figure}[t]
	\small
	\centering\includegraphics[width=70mm]{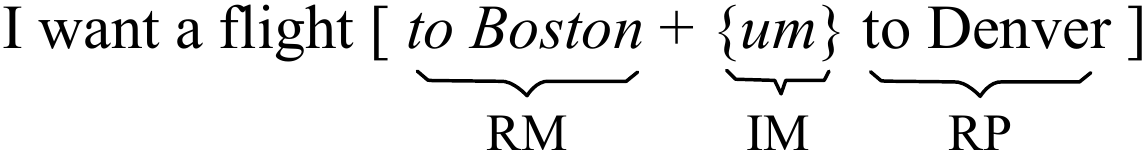}
	\caption{A sentence from the English Switchboard corpus with disfluencies annotated. RM=Reparandum, IM=Interregnum, RP=Repair. The preceding RM is corrected by the following RP. }
	\label{fig:example}
\end{figure}

Automatic speech recognition (ASR) outputs often contain various disfluencies, which create barriers to subsequent text processing tasks like parsing, machine translation, and summarization.
Disfluency detection~\cite{zayats2016disfluency,wang-che-liu:2016:COLING,wu-EtAl:2015:ACL-IJCNLP} focuses on recognizing the disfluencies from ASR outputs. 
As shown in Figure \ref{fig:example}, a standard annotation of the disfluency structure indicates the reparandum (words that the speaker intends to discard), the interruption point (denoted as `+', marking the end of the reparandum), an optional interregnum (filled pauses, discourse cue words, etc.) and the associated repair~\cite{shriberg1994preliminaries}.

\begin{table}[t]\small
	\setlength{\tabcolsep}{5pt}
	
	\begin{center}
		\renewcommand{\arraystretch}{1.1}
		\begin{tabular}{|l|c|}
			\hline
			\bf    Type &  \bf Annotation \\
			\hline
			repair &  [ I just + I ] enjoy working\\ 
			\hline
			repair &  [ we want + \{well\} in our area we want ] to\\ 
			\hline
			repetition  &  [it's + \{uh\} it's ] almost like\\
			\hline
			restart & [ we would like + ] let's go to the\\
			\hline
		\end{tabular}
	\end{center}
	\caption{ Different types of disfluencies.}
	\label{disf_example}
\end{table}

Ignoring the interregnum, disfluencies are categorized into three types: restarts, repetitions and corrections. 
Table \ref{disf_example} gives a few examples. 
Interregnums are relatively easier to detect as they are often fixed phrases, e.g. ``uh'', ``you know". 
On the other hand, reparandums are more difficult to detect in that they are in free form. 
As a result, most previous disfluency detection work focuses on detecting reparandums.

\begin{figure}[t]
	\small
	\centering\includegraphics[width=75mm]{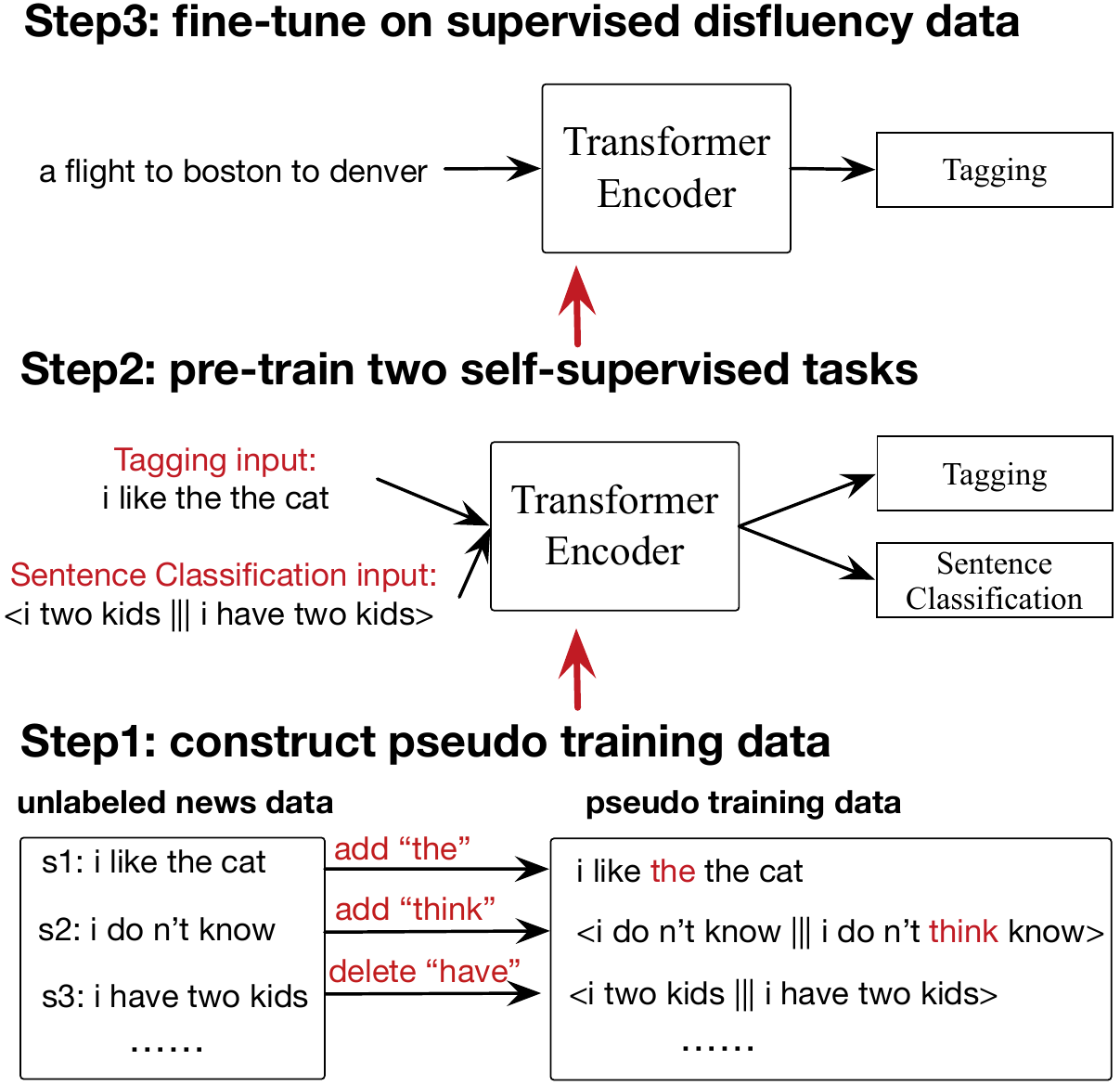}
	\caption{Illustration of our proposed methods. }
	\label{network}
\end{figure}

Most work~\cite{zayats2018robust,lou2018disfluency,wang2017transition,jamshid-lou-etal-2018-disfluency,zayats2019giving} on disfluency detection heavily relies on human-annotated data, which is scarce and expensive to obtain in practice. 
In this paper, we investigate self-supervised learning method ~\cite{agrawal2015learning,fernando2017self} to tackle this training data bottleneck.
Self-supervised learning aims to train a network on auxiliary tasks where ground-truth is obtained automatically.
The merits of this line work are that they do not  require manually annotations but still utilize supervised learning by inferring supervisory signals from the data structure.
Neural networks pre-trained with these tasks can be fine-tuned to perform well on standard supervised task with less manually-labeled data than networks which are initialized randomly. 
In natural language processing domain, self-supervised research mainly focus on word embedding \cite{mikolov2013efficient,mikolov2013distributed} or language model learning \cite{bengio2003neural,peters-etal-2018-deep,radford2018improving,devlin2018bert}.
Motivated by the success of self-supervised learning, we propose two self-supervised tasks for disfluency detection task, as shown in Figure \ref{network}.

The first task aims to tag corrupted parts from a disfluent sentence, generated by randomly adding words to a fluent sentence. 
Although there are discrepancies between the distribution of gold disfluency detection data and the generated sentences, this task endows the model to recover the fluent sentences from the disfluent ones, which matches the final goal of disfluency detection. 

The second task is sentence classification to distinguish original sentences from corrupted sentences.
We align each original sentence from the news dataset and another disfluent sentence generated by randomly deleting or adding words to the original fluent sentence. 
The goal of the task is to take these sentence pairs as input and predict which sentence is the fluent one. 
This task enables the model to distinguish grammatically-correct sentences from grammatically-incorrect ones. 
We hypothesize that this task is helpful for disfluency detection, as one core challenge for disfluency detection is to keep the output sentences grammatically-correct.

The second task can help the first by modeling sentence-level grammatical information.
Inspired by the hypothesis,
we combine these two tasks to jointly train a network based on the auto-constructed pseudo training data. 
The pre-trained network is later fine-tuned using  human-annotated  disfluency detection data.

Our contributions can be summarized as follows:
\begin{itemize}
	\item We propose two self-supervised tasks for disfluency detection to tackle the training data bottleneck. To our best knowledge, this is the first work to investigate self-supervised representation learning in disfluency detection.
	\item Based on the two self-supervised tasks, we further investigate multi-task methods for combining the two self-supervised tasks. 
	\item Experimental results on the commonly used English Switchboard test set show that our approach can achieve competitive performance compared to the previous systems (trained using the full dataset) by using less than 1\% (1000 sentences) of the training data. Our method trained on the full dataset significantly outperforms previous methods, reducing the error by 21\% on English Switchboard.
\end{itemize}

\section{Related Work}

\subsection*{Disfluency Detection}

Most work on disfluency detection focus on supervised learning methods, which mainly fall into three main categories: sequence tagging, noisy-channel, and parsing-based approaches. 
Sequence tagging approaches label words as fluent or disfluent using a variety of different techniques, including conditional random fields (CRF) ~\cite{georgila2009using,ostendorf2013sequential,zayats2014multi}
, Max-Margin Markov Networks (M$^3$N) \cite{qian2013disfluency}, Semi- Markov CRF \cite{ferguson-durrett-klein:2015:NAACL-HLT}, and recurrent neural networks~\cite{hough2015recurrent,zayats2016disfluency,wang-che-liu:2016:COLING}. 
Noisy channel models ~\cite{charniak2001edit,johnson2004tag,zwarts2010detecting,lou2018disfluency} use the similarity between reparandum and repair as an indicator of disfluency.
Parsing-based approaches \cite{rasooli2013joint,honnibal2014joint,wu-EtAl:2015:ACL-IJCNLP,yoshikawa2016joint} jointly perform dependency parsing and disfluency detection.
The joint models can capture long-range dependency of disfluencies as well as chunk-level information.

There exist a limited effort to tackle the training data bottleneck.
\citet{C18-1299} and \citet{dong2019adapting}  use an autoencoder method to help for disfluency detection by jointly training the autoencoder model and disfluency detection model.
They construct large-scale pseudo disfluent sentences by using some simple rules and use autoencoder to reconstruct the disfluent sentence.
We take inspiration from their method when generating disfluent sentences.
They achieve higher performance by introducing pseudo training sentence.
However, the performance of their method still heavily relies on annotated data.

\subsection*{Self-Supervised Representation Learning}

Self-supervised learning aims to train a network on an auxiliary task where ground-truth is obtained automatically.
Over the last few years, many self-supervised tasks have been introduced in image processing domain, which make use of non-visual signals, intrinsically correlated to the image, as a form to supervise visual feature learning \cite{agrawal2015learning,wang2015unsupervised}.

In natural language processing domain,  self-supervised research mainly focus on word embedding \cite{mikolov2013efficient,mikolov2013distributed} and language model learning \cite{bengio2003neural,peters-etal-2018-deep,radford2018improving}.
For word embedding learning, the idea is to train a model that maps each word to a feature vector, such that it is easy to predict the words in the context given the vector. 
This converts an apparently unsupervised problem  into a ``self-supervised" one: learning a function from a given word to the words surrounding it. 

Language model pre-training \cite{bengio2003neural,peters-etal-2018-deep,radford2018improving,devlin2018bert} is another line of self-supervised learning task. 
A trained language model  learns a function to predict the likelihood of occurrence of a word based on the  surrounding sequence of words used in the text. 
There are mainly two existing strategies for applying pre-trained language representations to down-stream tasks: feature-based and fine-tuning. 
The feature-based approach, such as ELMo \cite{peters-etal-2018-deep}, uses task-specific architectures that include the pre-trained representations as additional features. 
The fine-tuning approach, such as the Generative Pre-trained Transformer (OpenAI GPT)  \cite{radford2018improving} and BERT \cite{devlin2018bert}, introduces minimal task-specific parameters and is trained on the downstream tasks by simply fine-tuning the pre-trained parameters.

\citet{liu2016generating} propose a method to automatically generate large-scale pseudo training data for zero
pronoun resolution, and  propose a two-step training mechanism to overcome the gap between the pseudo training data and the real one.

Motivated by the success of self-supervised learning,  we propose a token-level tagging task and a  sentence-level classification task especially  powerful for disfluency detection task. 

\subsection*{Multi-Task Learning}

MTL (Multi-Task Learning) has been used for a variety of NLP tasks including named entity recognition and semantic labeling \cite{martinez-alonso-plank-2017-multitask}, super-tagging and chunking \cite{bingel-sogaard-2017-identifying} and semantic dependency parsing \cite{peng-etal-2017-deep}.
The benefits of MTL largely depend on the properties of the tasks at hand, such as the skewness of the data distribution \cite{martinez-alonso-plank-2017-multitask}, the learning pattern of the auxiliary and main tasks where ``target tasks that quickly plateau" benefit most from ``non-plateauing auxiliary tasks" \cite{bingel-sogaard-2017-identifying} and the ``structural similarity" between the tasks \cite{peng-etal-2017-deep}.
In our work, we use the sentence classification task to help the tagging task by integrating sentence-level grammatical information.

\section{Proposed Approach}

\subsection{Self-Supervised Learning Task}
Let $S = \{w_1, w_2, . . . , w_n\}$ be an ordered sequence of $n$ tokens, which is taken from raw unlabeled news data, assumed to be fluent. 
We then propose two self-supervised tasks.

\subsubsection*{Tagging Task}
The input of the tagging task is a disfluent sentence $S_{disf}$, generated by randomly adding words to a fluent sentence. 
$S_{disf}$ is fed into a transformer encoder  network to learn the representation of each word, $\{h_1, h_2, . . . , h_n\}$.
The goal is to detect the added noisy words by associating a label for each word, where the labels $D$ and $O$ means that the word is an added word and a fluent word, respectively.
Although the distribution of the tagging task data is different from the distribution of the gold disfluency detection data,  the training goal is to keep the generated sentences fluent by deleting disfluent words, which matches the goal of disfluency detection. 
We argue that the tagging model can capture more sentence structural information which is helpful for disfluency detection.

We start from a fluent sequence $S$ and introduce random perturbations to generate a disfluent sentence $S_{disf}$.
More specifically, we propose two types of perturbations:
\begin{itemize}
	\item \textit{Repetition$(k)$} : the $m$ (randomly selected from $one$ to $six$) words starting from the position $k$ are repeated.
	\item \textit{Inserting$(k)$} : we randomly pick a $m$-gram ($m$ is randomly selected from $one$ to $six$) from the news corpus and insert it to the position $k$.
\end{itemize}
For the input fluent sentence, we randomly choose $one$ to $three$ positions, and then randomly take one of the two perturbations for each selected position to generate the disfluent sentence $S_{disf}$.
It is important to note that it is possible that in some cases $S_{disf}$ will itself form a fluent sentence and hence violate the definition of the disfluent sentence.
We do not address this issue and assume that such cases will be relatively few and will not harm the training goal when the training data is large.

\subsubsection*{Sentence Classification Task}

The input of sentence classification task is a sentence pair $<S_1, S_2>$, where one is a fluent sentence and the other one is disfluent, generated by randomly adding or deleting some words from the corresponding fluent sentence. 
The sentence pair is fed into a transformer encoder network  to obtain a sentence pair representation $h_{s}$.
The goal of the task  is to discriminate between fluent sentence and corresponding disfluent one.
We define a label set, $\{add_0, add_1, del_0, del_1\}$, where $add_0$ and $del_0$ mean that the first input sentence $S_1$ is generated by randomly adding and deleting some words from the second sentence $S_2$, respectively.
We hypothesize that this task can capture sentence-level grammatical information, which is helpful for disfluency detection whose training goal is to keep the generated sentence fluent by deleting the disfluent words.

We construct two kinds of disfluent sentences for this task.
We use the same method described in the tagging task to construct the disfluent sentence $S_{add}$ with added noisy words.
For the disfluent sentence $S_{del}$ with deleted words, we consider a new type of perturbations:
\begin{itemize}
	\item \textit{Delete$(k)$} : for selected position $k$, $m$ (randomly selected from $one$ to $six$)  words starting from this position are deleted.
\end{itemize}
For the input fluent sentence, we randomly choose one to three positions, and then take the \textit{Delete$(k)$}  perturbation to generate $S_{del}$.
Note that one sentence can only be used to generate one kind of disfluent sentence to prevent the model from learning some statistical rules (e.g. the sentence with intermediate length is a fluent sentence) beyond our goals.

\subsection{Network Structure}

\begin{figure}[t]
	\small
	\centering\includegraphics[width=65mm]{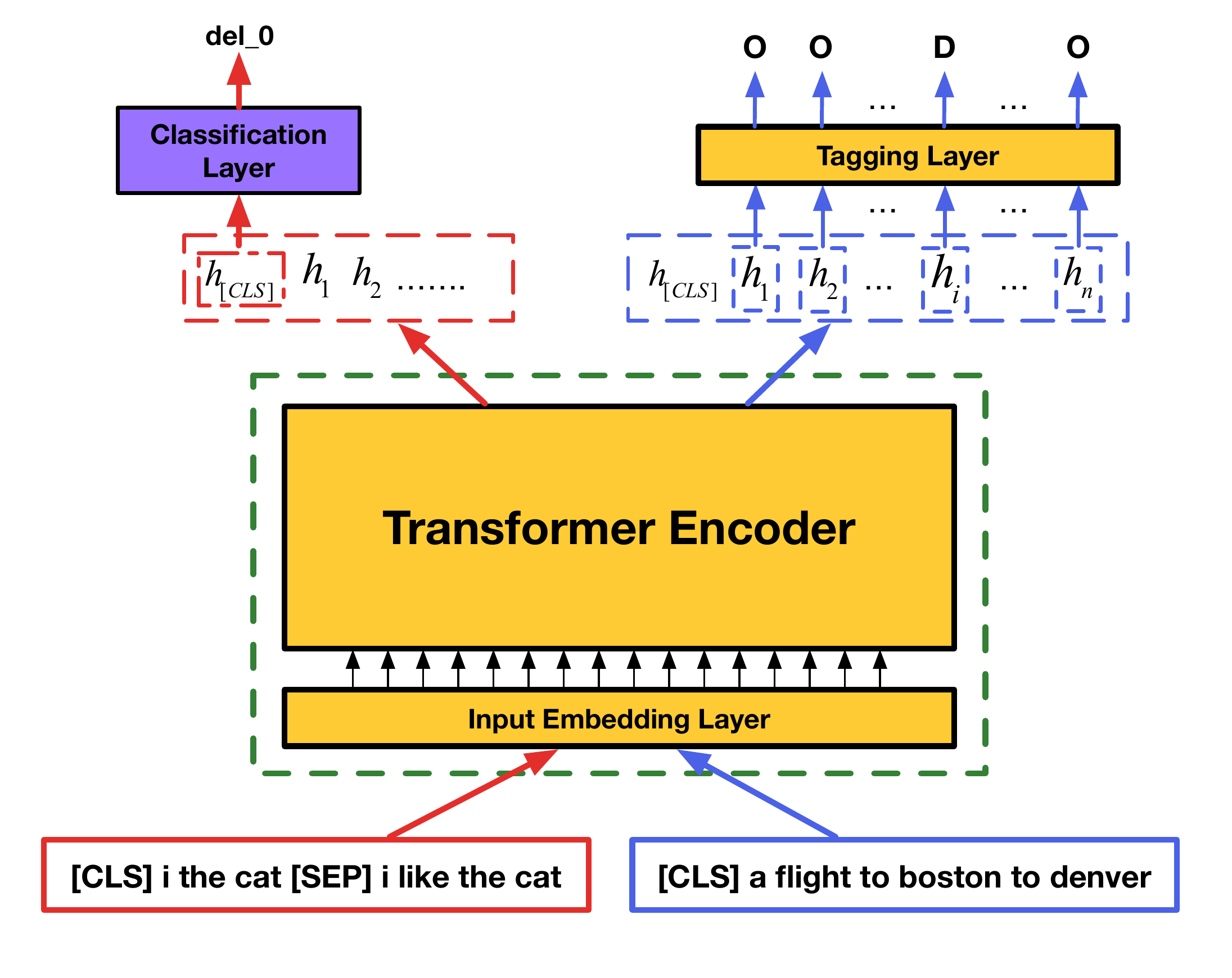}
	\caption{Model structure. The parameters of input embedding layer $I$, encoder layer $E$, and tagging layer $T$ (yellow box) are shared among pre-training and fine-tuning}
	\label{network1}
\end{figure}

As shown in Figure \ref{network1}, the model consists of four parts:  an input embedding layer $I$, an encoder layer $E$, a tagging layer $T$ for the tagging task, and a classification layer $C$ for the classification task.

For  $I$,   given a token, its input representation is constructed by summing the corresponding token, position, and segment embeddings.
For  $E$,  we use the multi-layer bidirectional transformer encoder described in  \citet{vaswani2017attention}.

For the tagging task, $E$ takes a sequence of input $([CLS], x_1, x_2, ..., x_n)$ and returns a representation sequence $(h_{[CLS]}, h_1, h_2, ..., h_n)$. 
Then the representation sequence $(h_1, h_2, ..., h_n)$ is sent to  $T$ to get a sequence of labels $(y_1, y_2, ..., y_n)$, where $y_i \in\{O, D\}$.

For the  sentence classification task,  $E$ takes two sequences of input $([CLS], x^1_1, x^1_2, ..., x^1_m, [SEP],  x^2_1, x^2_2, ...,  \\x^2_n)$ and returns a representation sequence $(h_{[CLS]}, h^1_1, h^1_2,  \\..., h^1_m, h_{[SEP]},  h^2_1, h^2_2, ..., h^2_n)$. 
Then we send the representation $h_{[CLS]}$ to   $C$ to get the classification label $y$, where $y \in\{add_0, add_1, del_0, del_1\}$.

\subsection{Multi-Task Pre-training Procedure}

Multi-task learning helps in sharing information between different tasks and across domains.
Our primary aim is to use the sentence classification task to help the tagging task by integrating sentence-level grammatical information.

Under the multi-task learning framework, the parameters of  $I$ and $E$ are shared.
We denote  $T_{ind}$ and $C_{ind}$ as the representations of  the tagging layer and the classification layer, respectively.
The total loss of the multi-task neural network is calculated as:
\begin{equation}
Loss = Loss_{tag} + Loss_{cl},\nonumber
\end{equation}
where $Loss_{tag}$ means the loss of tagging task, and $Loss_{cl}$ means the loss of sentence classification task.

In practice, we construct mini-batches of training examples, where 30\% of the data are single sentences used for the tagging task, and another 70\%  are sentence pairs for the sentence classification task.
Since parts of the encoder are shared among both tasks, we optimize both loss terms concurrently.

\subsection{Disfluency Detection Fine-tuning}
We directly fine-tune the pre-trained tagging model (including input embedding layer $I$,  encoder layer $E$, and tagging layer $T$) on gold human-annotated disfluency detection data.
Given a pre-trained  tagging model,  this stage converges faster as it only needs to adapt to the idiosyncrasies of the target disfluency detection data, and it allows us to train a robust disfluency detection model even for small datasets.
For fine-tuning, most model hyperparameters are the same as in pre-training, with the exception of the batch size, and number of training epochs.

\section{Experiment}

\subsection{Settings}
\textbf{Dataset}.\quad 
English Switchboard (SWBD) \cite{godfrey1992switchboard} is the standard and largest ($1.73 \times 10^5$ sentences for training ) corpus used for disfluency detection. 
We use  English Switchboard as  main  data.
Following the experiment settings in \citet{charniak2001edit}, we split the Switchboard corpus into train, dev and test set as follows: train data consists of all sw[23]$*$.dff files, dev data consists of all sw4[5-9]$*$.dff files and test data consists of all sw4[0-1]$*$.dff files.
Following \citet{honnibal2014joint}, we lower-case the text and remove all  punctuations and partial words.\footnote{words are recognized as partial words if they are tagged as `XX' or end with `-'.}
We also discard the `um' and `uh' tokens and merge `you know' and `i mean' into single tokens.

Unlabeled sentences are randomly extracted from WMT2017 monolingual language model training data (News Crawl: articles from 2016), consisting of English news\footnote{http://www.statmt.org/wmt17/translation-task.html}.
Then we use the methods  described previously  to construct the pre-training dataset.
The training set of the tagging task contains 3 million sentences, in which half of them are  pseudo  disfluent sentences $S_{disf}$ and others are fluent sentences directly extracted from the news corpus.
We use 9 million sentence pairs for the sentence classification task.

\noindent	\textbf{Metric}.\quad Following previous works \cite{ferguson-durrett-klein:2015:NAACL-HLT}, token-based precision (P), recall (R), and (F1) are used as the evaluation metrics.

\noindent \textbf{Baseline}.\quad 
We build two baseline systems including:(1) \textbf{Transition-based} \cite{wang2017transition} is a neural transition-based model  and achieves the current state-of-the-art result by integrating complicated hand-crafted features. 
We directly use the code released by \citet{wang2017transition}.\footnote{https://github.com/hitwsl/transition\_disfluency}
(2) \textbf{Transformer-based} is a multi-layer bidirectional transformer encoder with random initialization to directly train on human-annotated disfleuncy detection data. The network structure is similar to our model except for the classification layer $C$ in Figure \ref{network1}. We add it to show that the improvements do not come from the multi-layer bidirectional transformer encoder, yet from the pre-training process.

\begin{table*}[t]\small
	\setlength{\tabcolsep}{7pt}
	\begin{center}
		\renewcommand{\arraystretch}{1.2}
		\begin{tabular}{l|ccc|ccc|ccc|ccc}
			\hline
			\multirow{3}{*}{Method} & \multicolumn{6}{c|}{Full} & \multicolumn{6}{c}{1000 sents} \\
			\cline{2-7}\cline{8-13}
			& \multicolumn{3}{c|}{Dev} & \multicolumn{3}{c|}{Test} & \multicolumn{3}{c|}{Dev} & \multicolumn{3}{c}{Test} \\
			\cline{2-4}\cline{5-7}\cline{8-10}\cline{11-13}
			& P & R & F1 & P & R & F1 & P & R & F1& P & R & F1\\
			\hline
			Transition-based  &  92.2 & 84.7 & 88.3 &   92.1 & 84.1 & 87.9 &  82.2 &57.4 & 67.6& 81.2 & 56.7& 66.8\\
			\hline
			Transformer-based & 86.5 &70.4 &77.6& 86.1&	71.5&78.1&78.2&	51.3&62.0&79.1&51.1&62.1\\
			\hline
			Our self-supervised &92.9&	88.1&\textbf{90.4}&93.4&87.3	&\textbf{90.2}&90.0&82.8&\textbf{86.3}&88.6&83.7&\textbf{86.1}\\
			\hline
		\end{tabular}
	\end{center}
	\caption{ Experiment  results on English Switchboard data, where ``Full" means the results using 100\% human-annotated data, and ``1000 sents" means the results using less than 1\% (1000 sentences) human-annotated data.  }
	\label{main result}
\end{table*}

\begin{table}[t]\small
	\setlength{\tabcolsep}{5pt}
	
	\begin{center}
		\renewcommand{\arraystretch}{1.1}
		\begin{tabular}{l|ccc}
			\hline
			\bf    Method & \bf P & \bf R & \bf F1 \\
			\hline
			UBT \cite{wu-EtAl:2015:ACL-IJCNLP} &  90.3 & 80.5 & 85.1\\ 
			Semi-CRF (Ferguson et al., 2015) & 90.0 & 81.2 & 85.4\\
			Bi-LSTM (Zayats et al., 2016) &  91.8 & 80.6 & 85.9\\ 
			LSTM-NCM \cite{lou2018disfluency}  &  - & - & 86.8\\
			Transition-based \cite{wang2017transition}  &  91.1 & 84.1 & 87.5\\ 
			\hline
			Our self-supervised (1000 sents)  &  88.6 & 83.7 & 86.1\\ 
			Our self-supervised (Full) &  \textbf{93.4} & \textbf{87.3} & \textbf{90.2}\\ 
			\hline
			
		\end{tabular}
	\end{center}
	\caption{ Comparison with  previous state-of-the-art methods on the test set of English Switchboard. ``Full" means using 100\% human-annotated data  for fine-tuning, and ``1000 sents" means using less than 1\% (1000 sentences) human-annotated data   for fine-tuning.}
	\label{compare-previous-work}
\end{table}

\subsection{Training Details}

In all experiments including the transformer-based baseline and our self-supervised method, we use a transformer architecture with 512 hidden units, 8 heads,  6 hidden layers, GELU activations \cite{hendrycks2016bridging}, and a dropout of 0.1. 
We train our models with the Adam optimizer.

For the joint tagging and sentence classification objectives, we use streams of 128 tokens and a mini-batches of size 256.
We use learning rate of 1e-4 and epoch of 30.
When fine-tuning on gold disfluency detection data, most model hyperparameters are the same as in pre-training, with the exception of the batch size, learning rate, and number of training epochs.
We use batch size of 32, learning rate of 1e-5, and epoch of 20.

\subsection{ Performance On English Switchboard }

Table \ref{main result} shows the overall performances of our model on both development and test sets.
We can see that our self-supervised method outperforms the baseline methods in all the settings.
Surprisingly, our self-supervised method achieves almost 20  point improvements over transition-based method 	 when using less than 1\% (1000 sentences)  human-annotated disfluency detection data. 

We compare our  self-supervised model to five top performing systems, which rely on large-scale human-annotated data and complicated hand-crafted features. 
Our model outperforms the state-of-the-art, achieving a 90.2\%  F1-score as shown in Table \ref{compare-previous-work}. 
We  attribute the success to the strong ability to learn global sentence-level structural information.
Surprisingly, with less than 1\% (1000 sentences)  human-annotated training data, our model achieves comparable F1-score as the previous top performing systems using 100\% human-annotated training data, which shows that our self-supervised method can substantially reduce the need for human-annotated training data.  
Note that we do not compare our work with the work of \citet{C18-1299}  using semi-supervised method for disfluency detection.
\citet{C18-1299} treat interregnum and reparandum types equally when training and evaluating their model, while others (including ours and all the baselines in Table \ref{compare-previous-work}) only focus on reparandums which are more difficult to detect.

\begin{table}[t]
	\setlength{\tabcolsep}{5pt}
	\centering
	\small
	\renewcommand{\arraystretch}{1.1}
	\begin{tabular}{l|ccc|ccc}
		\hline
		\multirow{2}{*}{Method} & \multicolumn{3}{c|}{Full} & \multicolumn{3}{c}{1000 sents} \\
		\cline{2-4}\cline{5-7}
		& P & R & F1 & P & R & F1 \\
		\hline
		Random-Initial &86.1&71.5&78.1&79.1&51.1&62.1\\
		\hline
		Tagging &91.8&84.0&87.7&85.1&79.6&82.3\\
		\hline
		Classification&91.2&83.1&87.0&83.2&78.3&80.7\\
		\hline
		Multi-Task &\textbf{93.4}&\textbf{87.3}&\textbf{90.2}&\textbf{88.6}&\textbf{83.7}&\textbf{86.1}\\
		\hline
	\end{tabular}
	\caption{  Ablation over the two self-supervised tasks. ``Random-Initial" means training transformer network on gold disfluency detection data with random initialization.}
	\label{multi-task}
\end{table}

\begin{table}[t]
	\setlength{\tabcolsep}{10pt}
	\centering
	\small
	\renewcommand{\arraystretch}{1.1}
	\begin{tabular}{ccc | cc}
		\hline
		\multicolumn{3}{c | }{Hyperparams} & \multicolumn{2}{c}{Dev Set Performance} \\
		\hline
		\#L & \#H & \#A  & F1 (1000 sents) &  F1 (Full)\\
		\hline
		2&256&4&  78.1& 85.7\\
		\hline
		2&256&6&  79.7& 86.8\\
		\hline
		4&256&6& 81.8& 88.2\\
		\hline
		4&256&8& 82.9& 89.1\\
		\hline
		6&256&8& 84.5 & 89.6\\
		\hline
		\textbf{6}&512& 8 & \textbf{86.3}&\textbf{90.3}\\
		\hline
	\end{tabular}
	\caption{ Ablation over model size. \#L = the number of layers; \#H = hidden size; \#A = number of attention heads. “Full” means fine-tuning on  100\%  gold training data, and “1000 sents” means fine-tuning on  less than 1\% (1000 sentences) gold training data.}
	\label{model-size}
\end{table}

\begin{figure*}[!th]
	\centering
	\begin{minipage}{0.32\linewidth}\centering
		\includegraphics[width=5.0cm]{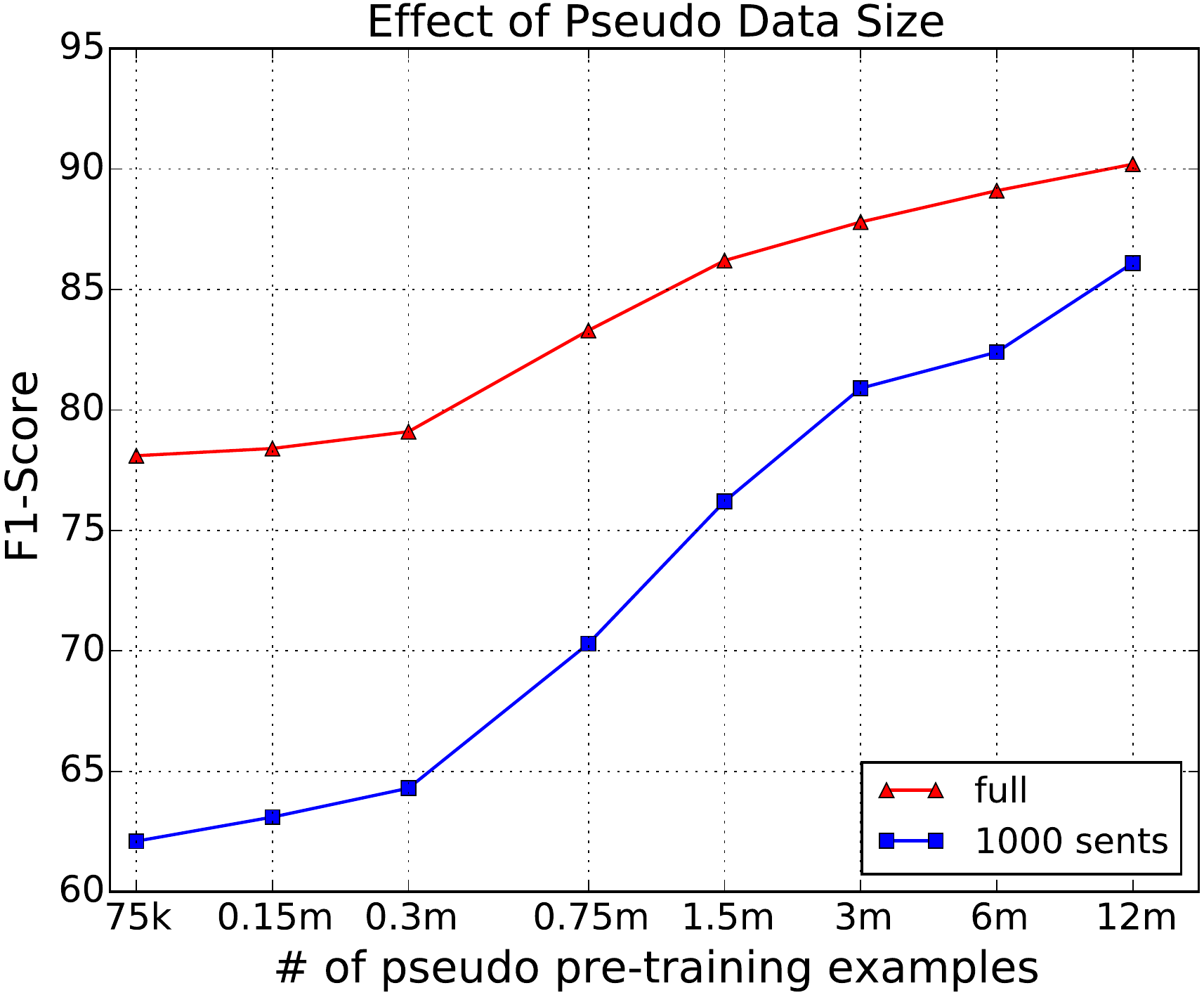}\\ 
		(a) 
	\end{minipage}
	\begin{minipage}{0.32\linewidth}\centering
		\includegraphics[width=5.0cm]{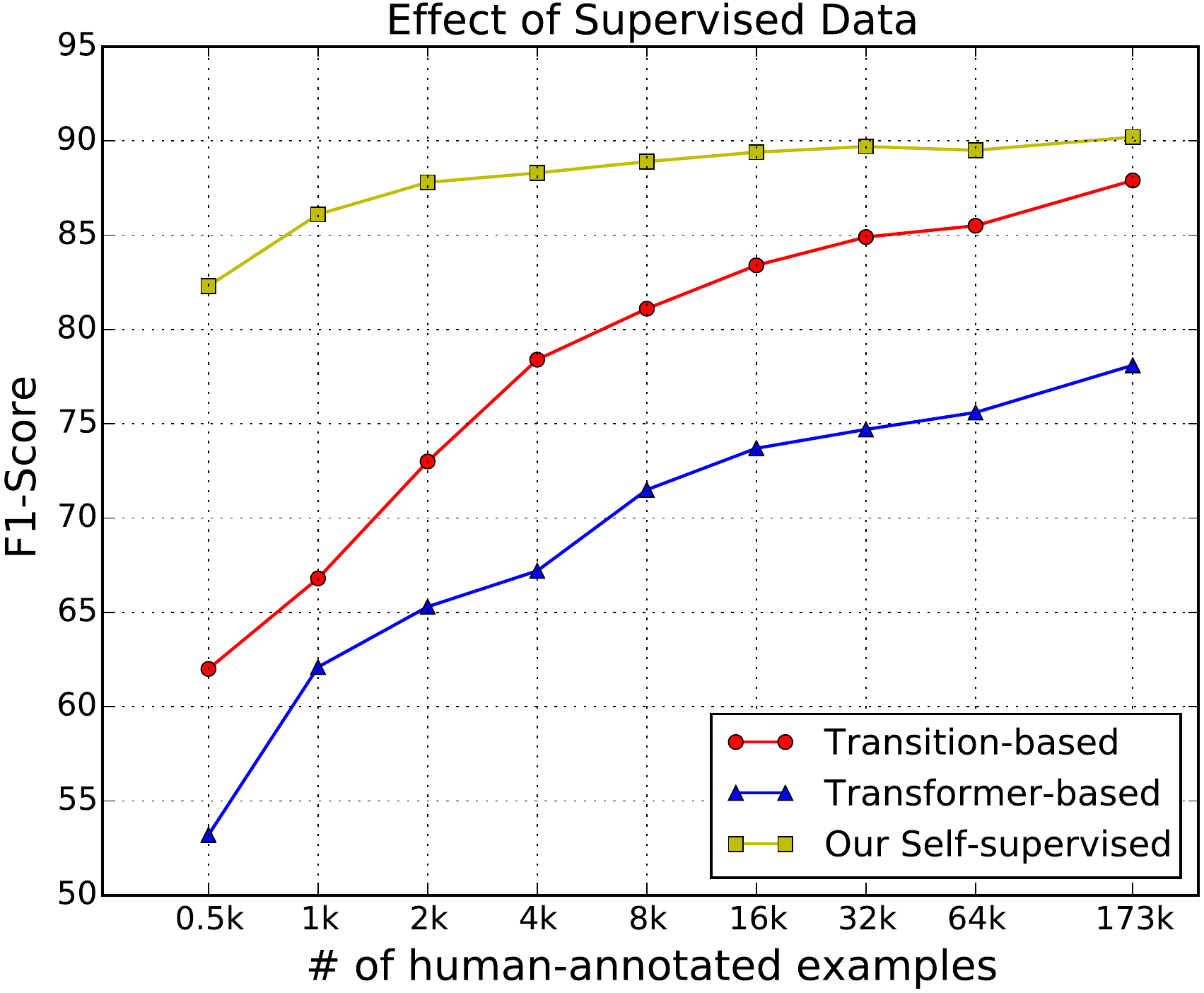}\\ 
		(b) 
	\end{minipage}
	\begin{minipage}{0.32\linewidth}\centering
		\includegraphics[width=5.0cm]{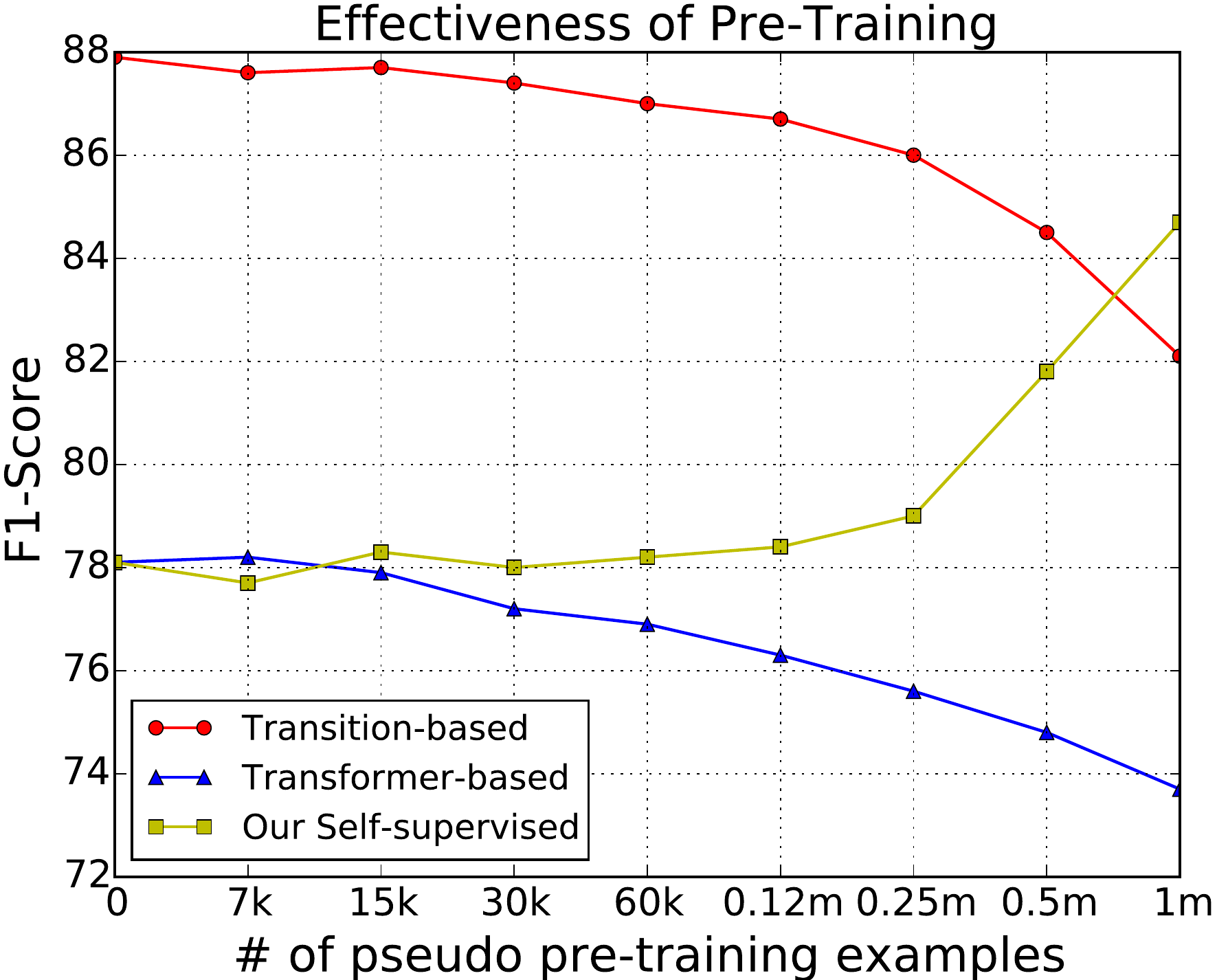}\\ 
		(c) 
	\end{minipage}
	\begin{minipage}{0.32\linewidth}\centering
		\includegraphics[width=5.0cm]{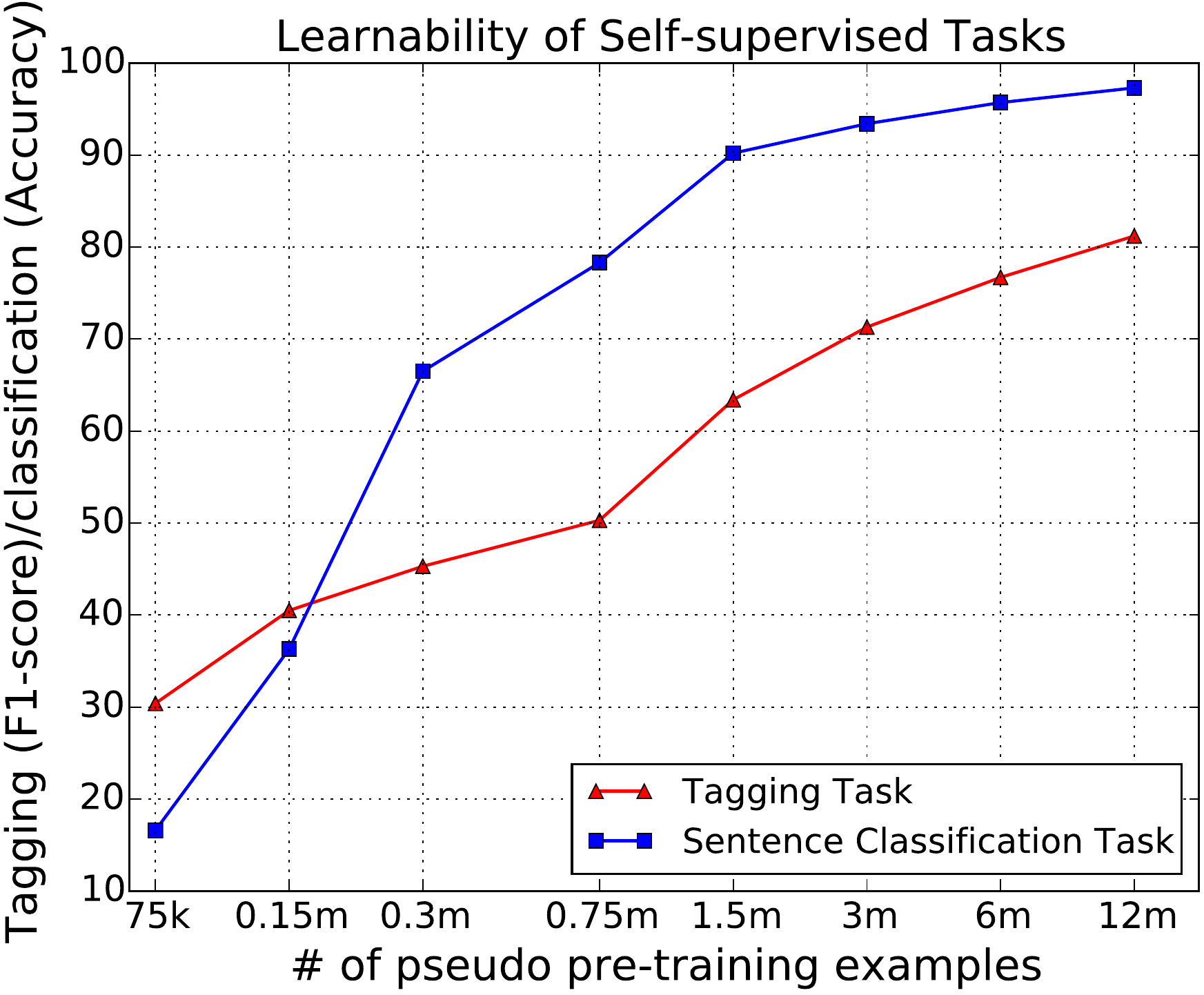}\\
		(d) 
	\end{minipage}
	\begin{minipage}{0.32\linewidth}\centering
		\includegraphics[width=5.0cm]{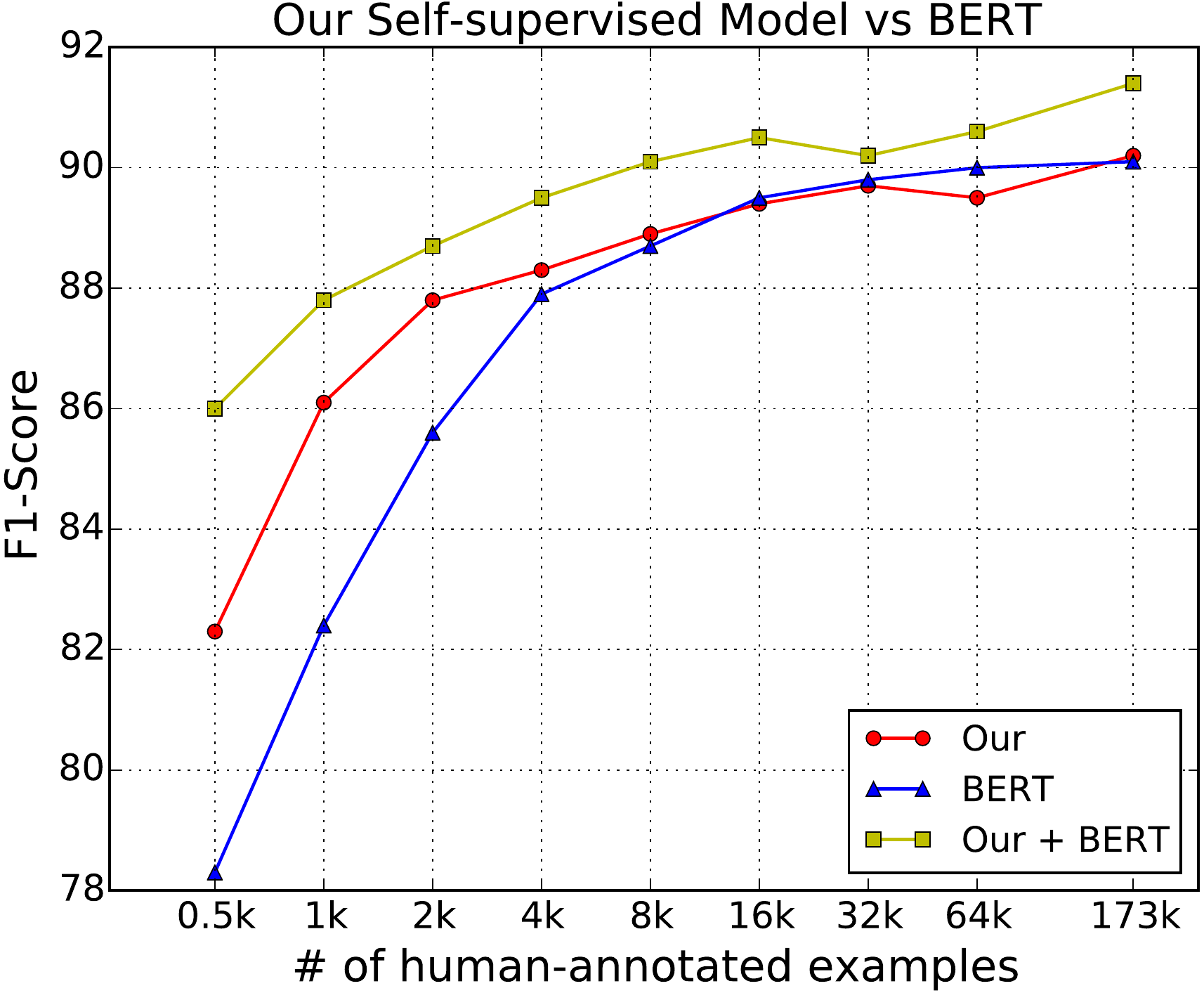}\\
		(e) 
	\end{minipage}
	
	\caption{(\textbf{a}) Plot showing the impact of pseudo training data size to disfluency detection. (\textbf{b}) Plot showing the impact of human-annotated data size when fine-tuning. (\textbf{c}) Plot showing the effectiveness of pre-training compared with the baseline methods of directly merging  gold training data  with the  pseudo training data. (\textbf{d}) Plot showing learnability of self-supervised tasks. (\textbf{e}) Plot showing the effectiveness of multi-task learning.}
	\label{zhexian}
\end{figure*}

\section{ Ablation Studies }
\subsection{ Effect of the two Self-Supervised Tasks }
We explicitly compare the impact of the tagging task and the sentence classification task.
As shown in Table \ref{multi-task},  both of our two self-supervised tasks  achieve higher performance compared with the baseline system with random initialization.
Higher performance is achieved by combining the two self-supervised tasks, which demonstrates that the two tasks can coexist harmoniously,  and share useful information between each other.

\subsection{Effect of Model Size}
We explore the effect of model size on fine-tuning disfluency detection task. 
We mainly  tune  the number of  layers, hidden units, and attention heads, while otherwise using the same hyperparameters as described previously.
In order to avoid the impact of randomness, we run 3 random restarts of  fine-tuning and report the average F1-Score  on the Dev set.
Results are shown in Table \ref{model-size}.
We can see that larger models always  lead to a strict F1-score improvement on the Dev set.
It is also  surprising that  larger models always reduce the performance gap between fine-tuning using full gold training data and fine-tuning using less than 1\% (1000 sentences) training data.
We believe that much more performance will be gained if we keep increasing the model size.

\section{Analysis}

\subsection{Varying Amounts of Pseudo Data}
We observed the impact of pseudo training data size to disfluency detection task.
Figure ~\ref{zhexian} (a) reports the results of adding varying amounts of pseudo training data to the self-supervised pre-training model. 
We observe that  F1-score keeps growing when the amount of automatically-generated data increases.
We conjecture that our two self-supervised tasks and disfluency detection task can coexist harmoniously, and more automatically-generated training data will bring more structural information. 
Another surprising observation is that the performance on the small supervised  dataset (1000 sentences) grows faster, which shows that our method  has huge potential to tackle the training data bottleneck.

\subsection{Varying Amounts of Supervised Data}
We explore how fine-tuning  scales with human-annotated  data size, by varying the amount of human-annotated training  data the model has access to.
We plot F1-score  with respect to the amounts of human-annotated disfluency detection data for fine-tuning in Figure ~\ref{zhexian} (b).
Compared with the baseline systems, fine-tuning based on our self-supervised models improves performance considerably when limited gold human-annotated training data is available, but those gains diminish with more high-quality human-annotated data. 
Using only 2\% of the labeled data, our approach already performs as well or better than the previous state-of-the-art transition-based method using 100\% of the human-annotated training data, demonstrating that our self-supervised tasks are particularly useful on small datasets.

\subsection{ Effectiveness of Pre-Training}
We explore the contribution of pre-training to the final experimental results.
As the pseudo-training data ($S_{disf}$ constructed by adding words) is similar to the gold disfluency detection data in format, a natural idea is to directly train previous methods with the mixed pseudo $S_{disf}$ data and gold training data. 
So we re-train the baseline transition-based method and transformer-based method by merging  the gold training data  with the  pseudo $S_{disf}$ data.
As shown in Figure ~\ref{zhexian} (c),   F1-scores of the two baseline methods keep decreasing when the amount of the pseudo training data increases, while F1-score of our self-supervised method keeps increasing. 
The results show that our self-supervised method  is much more effective compared with the methods of directly merging  the gold training data  with the pseudo data.
We attribute the F1-score decrease of baseline methods to the discrepancies between the distribution of gold disfluency detection sentence and the pseudo disfluent sentence $S_{disf}$.

\subsection{Learnability of Self-Supervised Tasks}
Our self-supervised tasks are very similar to supervised tasks, excepted that the training data is collected without manual labeling.
To prove the  learnability of our self-supervised tasks,  we explore the performance of our self-supervised models on the  pseudo testing data when pre-training.
We plot the performance  with respect to the amounts of pseudo training data for pre-training in Figure \ref{zhexian} (d).
The performance keeps growing when the amount of automatically-generated data increases,  achieving about 80\% F1-score on tagging task and 97\% accuracy on sentence classification task, respectively.
The results show that our self-supervised tasks are reasonable, which can really capture more sentence structural information.

\subsection{Repetitions vs Non-repetitions}
Repetition disfluencies are  easier to detect and even some simple hand-crafted features can handle them well. 
Other  types of reparandums such as repair are more complex \cite{zayats2016disfluency,ostendorf2013sequential}.
In order to better understand model performances, we evaluate our model's  ability  to detect repetition vs. non-repetition (other) reparandum. 
The results are shown in Table \ref{repetion-test}. 
All three models achieve high scores on repetition reparandum. 
Our self-supervised model  is much better in predicting non-repetitions compared to the two baseline methods. 
We conjecture that our self-supervised tasks can capture more sentence-level structural information.

\subsection{Comparison with BERT}
We would like to see the performance comparison between our pre-trained model and BERT.
The large version of pre-trained BERT model (24-layer transformer blocks, 1024 hidden-size, and
16 self-attention heads, totally 340M parameters) is used for the comparison.
In our experiment, we follow the hyper-parameters (e.g. batch size of 32) of \citet{devlin2018bert}
when fine-tuning BERT. 
Additionally, for the large version of pre-trained BERT model, fine-tuning was sometimes unstable on small datasets,
so we  run 3 random restarts and select the best model on the development set.

Compared with BERT, we  use much smaller training corpus and model parameters (6-layer transformer blocks, 512 hidden-size, and 8 self-attention heads) limited by devices.
Results are shown in Table \ref{bert-init}.
Both our method and BERT outperforms the baseline model with random initialization, which proves the strong ability of pre-training model.
Although our pre-training corpus  and model parameters   are much smaller than BERT, we achieve a similar result with BERT  when fine-tuning on full gold training data.
Surprisingly, our method achieves 3.7 point improvements over BERT when fine-tuning on 1\%(1000 sentences) of the gold training data.
We also  try to combine our pre-train model and BERT by concatenating their hidden representation.
The result shows that much higher performance is achieved.
We also plot  the performance with respect to the length of human-annotated disfluency detection data  in Figure \ref{zhexian} (e). 
The performance is always much higher by combining our pre-train model and BERT.
This proves that our model and BERT can coexist harmoniously, and capture different aspects of information helpful for disfluency detection.

\begin{table}[t]\small
	\setlength{\tabcolsep}{7pt}
	
	\begin{center}
		\renewcommand{\arraystretch}{1.2}
		\begin{tabular}{l | c c c}
			\hline
			\bf    Method & \bf Repet & \bf Non-repet & \bf Either \\
			\hline
			Transition-based & 93.8	&68.3	&87.9\\ 
			\hline
			Transformer-based & 93.6&	58.9&	78.1\\ 
			\hline
			Our self-supervised  &93.7&	70.8&\textbf{90.2}\\
			\hline
		\end{tabular}
	\end{center}
	\caption{F1-score of different types of reparandums on English Switchboard test data.}
	\label{repetion-test}
\end{table}

\begin{table}[t]\small
	\setlength{\tabcolsep}{8pt}
	
	\begin{center}
		\renewcommand{\arraystretch}{1.2}
		\begin{tabular}{l|c c}
			\hline
			\bf    Method & \bf F1 (Full)  & \bf F1 (1000 sents)\\
			\hline
			Random-Initial & 78.1	&62.1\\ 
			\hline
			BERT-fine-tune & 90.1	&82.4\\ 
			\hline
			Our self-supervised & 90.2&	86.1\\ 
			\hline
			Combine  &\bf 91.4&\bf	87.8\\
			\hline
		\end{tabular}
	\end{center}
	\caption{ Comparison with BERT.  ``random-initial" means training transformer network on gold disfluency detection data with random initialization. ``combine" means concatenating  hidden representations of BERT and our self-supervised models for fine-tuning.}
	\label{bert-init}
\end{table}

\section{Conclusion}
In this work, we propose two self-supervised tasks to tackle the training data bottleneck. 
Experimental results on the commonly used English Switchboard test set show that our approach can achieve competitive  performance compared to the previous systems (trained using the full dataset) by using less than 1\% (1000 sentences) of the training data. 
Our method trained on the full dataset significantly outperforms previous methods, reducing the error by 21\% on English Switchboard.

\section{ Acknowledgments}
We thank the anonymous reviewers for their valuable comments.
Shaolei Wang  was supported by China Scholarship Council (CSC), and the National Natural Science Foundation of China (NSFC) via grant 61976072, 61632011 and 61772153. Wanxiang Che is the corresponding author.

\bibliography{AAAI-WangS.1634}
\bibliographystyle{aaai}

\end{document}